\documentclass[11pt]{article}
\usepackage{nodalida2025}
\usepackage{times}
\usepackage[hyphens]{url}
\usepackage{latexsym}

\usepackage{graphicx}
\usepackage{hyperref}

\aclfinalcopy 

\title{Finnish SQuAD: A Simple Approach to Machine Translation of Span Annotations}

\author{Emil Nuutinen, Iiro Rastas and Filip Ginter \\
  TurkuNLP, Department of Computing \\
  University of Turku, Finland \\
  \texttt{emenuu, iiro.t.rastas, figint@utu.fi}}

\date{}

\begin{document}
\maketitle
\begin{abstract}
We apply a simple method to machine translate datasets with span-level annotation using the DeepL MT service and its ability to translate formatted documents. Using this method, we produce a Finnish version of the SQuAD2.0 question answering dataset and train QA retriever models on this new dataset. We evaluate the quality of the dataset and more generally the MT  method through direct evaluation, indirect comparison to other similar datasets, a backtranslation experiment, as well as through the performance of downstream trained QA models. In all these evaluations, we find that the method of transfer is not only simple to use but produces consistently better translated data. Given its good performance on the SQuAD dataset, it is likely the method can be used to translate other similar span-annotated datasets for other tasks and languages as well. All code and data is available under an open license: data at HuggingFace TurkuNLP/squad\_v2\_fi, code on GitHub TurkuNLP/squad2-fi, and model at HuggingFace TurkuNLP/bert-base-finnish-cased-squad2.
\end{abstract}

\section{Introduction}\label{sec1}

Question answering (QA) is an important practical information retrieval task as well as a common benchmark of computational models of human language. Extractive QA models are typically built as a two step retriever-reader pipeline, first retrieving the documents relevant to the query (retriever) and then using an encoder model to extract the correct answer span from those documents (reader). Generative QA models replace the reader component with a generative large language model (LLM), in an approach commonly referred to as retrieval-augmented generation (RAG).

No matter which of the QA paradigms is applied, large-scale question answering datasets such as the SQuAD dataset play a key role. Both in terms of benchmarking model performance, and model training. Whereas for extractive QA these datasets are used directly, in generative LLM development, QA datasets are commonly used as a source of examples for instruction fine-tuning. Unfortunately, these large-scale QA datasets are mostly available only for English and a small number of well-resourced languages, making the direct development of retriever-reader QA models for languages without such a dataset almost impossible, as well as negatively impacting benchmarking of LLM-based QA. 

With the improvements to machine translation (MT) output quality seen in the recent years, machine translating datasets is becoming a frequent choice to obtain a dataset in a new language in cases where native annotation is not possible due to lack of resources. While such an approach is technically very simple to implement for datasets consisting of unannotated text, it becomes considerably more complex for datasets with dense text span annotations, such as the QA datasets. Numerous approaches have been introduced aiming to transfer the span annotations during translation. In this paper we contribute to this overall line of research by demonstrating a simple, yet effective approach to translate English question answering datasets to Finnish (or other languages) using a little-known feature of the DeepL machine translation service. 

The primary contribution of this paper is a Finnish version of the publicly available sections of the SQuAD~2.0 dataset. This dataset can serve both for the development of extractive QA systems on top of Finnish encoder models, as well as provide a source of Finnish data for instruction tuning and benchmarking of Finnish LLMs. Our other contribution is a Finnish extractive QA model trained on this dataset.

The paper is organized as follows: In Section 2 we review prior work on machine translating QA-datasets. In Section 3 we explain our process of machine translating these datasets. In Section 4 we evaluate the new resource and compare it to other similar resources. Finally, Section 5 concludes the work.

\section{Related Work}\label{sec2}\label{sec:relwork}

There are numerous open-domain question answering datasets for English. Among the most commonly used is the Stanford Question Answering Dataset (SQuAD). SQuAD1.1 ~\cite{rajpurkar_squad_2016} consists of 100,000 questions posed by crowdsourced workers on a set of text passages (paragraphs of 536 Wikipedia articles). The questions are produced by the workers, while the answers constitute spans present in the text passages. SQuAD2.0 ~\cite{rajpurkar_know_2018} is a superset of SQuAD1.1 with an additional 50,000 crowdsourced unanswerable distractor questions that only make the impression of being answered in the given passage.\footnote{Note, however, that the test set of the SQuAD datasets is kept private, and the publicly available data contains 98,169 question-answer pairs for SQuAD1.1 and 92,749 answerable plus 49,434 unanswerable questions in SQuAD2.0.} 
Native human generated question answering datasets for other languages include Chinese ~\cite{cui_span-extraction_2019}, Korean ~\cite{lim_korquad10_nodate} and French ~\cite{dhoffschmidt_fquad_2020}, but a large number of languages lack a large QA dataset. 

The SQuAD dataset has been machine translated to several languages. Arabic ~\cite{mozannar_neural_2019} SQuAD1.1 version starts by machine translating the passages, questions and answers separately. Subsequently, all the paragraphs and answers are transliterated to Arabic and the span of text of length at most 15 words with the least edit-distance with respect to the answer is identified. Only 231 articles containing 48,344 question-answer pairs are translated, and a full 25,490 question-answer pairs are not recovered by the initial translation and the transliteration heuristic step is applied. A reported small-scale evaluation shows that approximately 64\% of these are correctly recovered.

Persian ~\cite{abadani_parsquad_2021} SQuAD2.0 version starts by machine translating the passages, questions and answers separately. Then an alignment is established by finding the position of the sentence that the answer appears in the English dataset. If the translated answer does not appear in the equivalent translated sentence, the question-answer pair is removed from the final dataset. The final dataset salvages 70,560 question-answer pairs.

The TAR-method (Translate-Align-Retrieve) used to create the Spanish translation of SQuAD1.1 ~\cite{carrino_automatic_nodate} also starts by machine translating the passages, questions and answers separately. If the translated answer can be found in the translated passage, it is retrieved as is. In the opposite case, a word alignment between the source and translated passage is established using the \emph{eflomal} word alignment method ~\cite{ostling_efficient_2016} and this alignment is then used to locate the translated answer. The final dataset salvages almost all of the question-answer pairs, but a manual error analysis showed that 50\% of the answer spans were either misaligned (7\%) or under-/over-extended (43\%). 

For Finnish, which is our target language of interest, there exists an earlier machine translated version of the SQuAD2.0 dataset ~\cite{kylliainen_neural_2022,kylliainen_question_nodate}. The passages, questions and answers are translated separately and their spans in the translations are identified using a number of normalization steps designed to improve the chance of successful matching. The dataset preserves 66,000 question-answer pairs from the original approx.\ 92,000.

The unpublished Swedish translation of SQuAD2.0\footnote{\url{https://towardsdatascience.com/swedish-question-answering-with-bert-c856ccdcc337}} deviates from the common approach, and translates one question-answer pair at a time, marking the answer span with a recognizable token (e.g. ``[0]''), and retrieves the span after translation, relying on the MT system preserving the special tokens. This process is reported to preserve 90\% of the original question-answer pairs.


In a more recent approach, a separate alignment model is first trained for the target language ~\cite{masad-etal-2023-automatic}. Then each context, question, and answer are translated together as a single unit using the Google Translate service. If the answer is not found with exact matching from the translation, the alignment model is used. Finally, if the first two steps fail, the context is segmented into subsets of words with a total word count that approximates the word count of the answer. Then the embeddings of the answer and all the context segments are calculated using a pre-trained multilingual BERT model from which the closest segment to the answer is searched using cosine similarity with a threshold on the similarity score to prevent weak alignments. This method is reported to preserve 93.4\% of the original question-answer pairs.

In another recent approach, an annotated clinical corpus is translated from English to Dutch \cite{10.1093/jamia/ocae159}. In the dataset the annotation and the context are stored separately. In the paper the annotations are first integrated directly into the clinical text by enclosing the text span and the CUI (concept unique identifier) in square brackets `[[text span] [CUI]]`. Then the text with embedded annotations is machine translated, keeping the annotations intact. Finally the annotations are extracted from the translated text using regular expressions to separate the annotations and the context again to the original format. The Google Translate service and GPT 4 Turbo are compared. The Google Translate service lost up to 1.7\% of annotations and GPT4 Turbo lost up to 5.9\%. Most of lost annotations for Google were formatting errors, but for GPT, the lost annotations were mostly entirely omitted.

In summary, the clearly most common approach to machine translating datasets with span level annotations relies on translating the elements in isolation, and subsequently identifying through a varied set of heuristics their positions in the translated passages. This is naturally an error-prone process due to the fact that the answers when translated in isolation are not guaranteed to match their in-context translation within the passage, preventing reliable alignment. This is demonstrated by the substantial proportions of ``lost'' examples reported for most of these machine translated datasets. And while metadata-tagging approaches like that of ~\citet{10.1093/jamia/ocae159} preserve most of the annotations, they are not able to preserve overlapping annotations without multiple rounds of translations.


In the following, we apply an approach which uses the functionality of a commercial MT engine to avoid the tedious alignment of answer segments with the original passages.

\section{Methods and Data}\label{sec3}

\subsection{Markup-based Transfer}

To create a translated version of the SQuAD dataset (or any other extractive QA dataset for that matter), not only the questions and underlying text passages need to be translated, but also the answer spans need to be correctly identified. Further, since the QA datasets often have many question-answer pairs for each passage, the answer spans may partially overlap.

Our work is based on the DeepL commercial machine translation service\footnote{\url{https://www.deepl.com/translator}} which is very popular among users thanks to its excellent translation output quality, which has also been reported in numerical benchmarks (e.g. ~\citet{shaitarova-etal-2023-machine}). In particular, we capitalize on the simple observation that DeepL is capable of translating formatted documents. This feature is crucial for professional translators---the primary users of the service---who need to translate not only the text of the source documents, but also preserve their formatting. In practice, this means that the input of DeepL can be a textual document with formatting (a Word document) and the service produces its translated version with the formatting preserved. This, in turn, gives us the combination of a high-quality machine translation system, an obviously necessary condition for successful machine translation of training data, with the ability to link text spans between the source and target documents through formatting. We first utilized this property of DeepL  to machine translate a relation extraction dataset to a number of languages. In that work, the annotation did not exhibit overlapping spans ~\cite{bassignana-etal-2023-multi}.

\begin{figure*}[htbp]
\graphicspath{ {./figures/} }
\includegraphics[width=\textwidth]{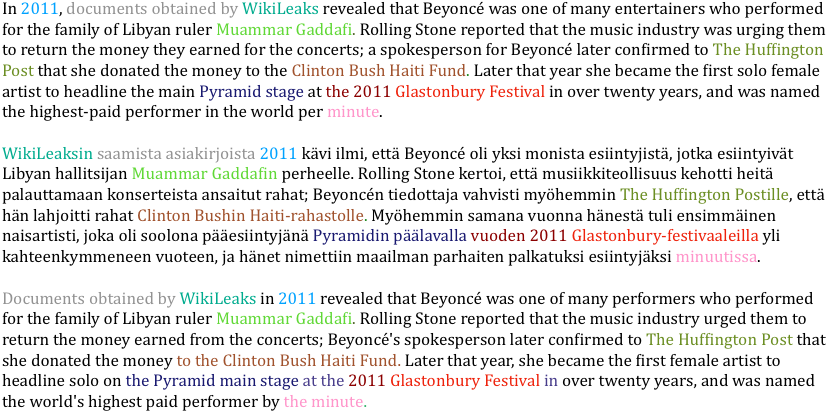} 
\caption{Example of the colored answer spans from an actual SQuAD passage: the original English passage (top), its Finnish translation (middle), and its backtranslation from Finnish into English (bottom). This example is shown as-is without any manual corrections (other than adjusting colors for better readability). Note the two overlapping answers \emph{documents obtained by WikiLeaks} and \emph{WikiLeaks} at the very beginning of the passage.}
\label{fig:colored-translation}
\end{figure*}

The answer spans in the dataset can be trivially encoded as colored text spans in the input documents, where the color uniquely differentiates the individual answer spans. This is somewhat complicated by the fact that the answers may overlap in the dataset. A simple solution is to consider the overlapping region to be a separate span, and assign it a distinct color, and reverse this mapping when reconstructing the dataset after translation. Another approach would have been, for instance, to translate each context several times for different non-overlapping subsets of entities. Nevertheless, having observed that in our case the former approach did not cause any clear degradation of the output, we chose to not pursue the latter approach, which would have increased the cost of translation\footnote{The overall translation cost of SQuAD was approximately $20$\texteuro.} and complexity of reconstructing the data. The translation process with formatting is illustrated on an actual example from the dataset in Figure~\ref{fig:colored-translation}.

Observing that oftentimes the answer spans were over-extended by a trailing punctuation symbol during translation, the only post-processing we apply is to strip from each translated span any trailing punctuation. This, in our view, has no negative impact on the QA task.

One aspect, common to all machine translation approaches to SQuAD irrespective of the method of annotation transfer, is that the answer spans in the original SQuAD data are always continuous, which is not necessarily the case in the translation simply due to the properties of the target language. In these cases, the translation system often correctly highlights the discontinuous regions in the translation, however the SQuAD data file format does not represent discontinuous answer regions, nor do the off-the-shelf model architectures developed for SQuAD allow for generation of discontinuous spans. To deal with this, and still allow the data to be used also with standard architectures, we include in the final data files both the original potentially discontinuous spans (as a separate key) and continuous spans obtained by simply spanning from the first to the last discontinuous span. In our dataset, only 2.6\% of the answers are discontinuous, many of which are translation artefacts upon manual inspection.

\subsection{Finnish SQuAD2.0}

We used the method described above to machine translate the publicly available sections of the SQuAD2.0 dataset to Finnish. The resulting dataset preserves 90,233 question-answer pairs from the original 92,749, i.e.\ 97.2\% of the dataset. This is substantially more than the majority of SQuAD machine translations discussed in Section~\ref{sec:relwork}

\subsection{Finnish Extractive QA Models}

We train an extractive QA model on the Finnish SQuAD dataset using the Finnish FinBERT-base model ~\cite{virtanen_finbert_2019} and the standard approach to span-detection with BERT models described by \newcite{devlin_bert_2019} and implemented in the Hugging Face Transformers library ~\cite{wolf_transformers_2020}.
Since the English SQuAD2.0 test set is not publicly available, we fine-tune our model using only the train set and use the validation set for evaluation. This matches how most of the other reported models are trained and evaluated.

Interestingly, state-of-the-art performance models for the English SQuAD dataset almost uniquely rely on the ALBERT pre-trained model ~\cite{lan_albert_2020}, with very substantial reported gains ~\cite{lan_albert_2020,abadani_parsquad_2021} over the standard BERT models. In order to test whether a similar effect can be obtained also for Finnish, we also pre-train a series of Finnish ALBERT models (FinALBERT) and fine-tune them on the Finnish SQuAD dataset.

The pretraining of FinALBERT follows the original ALBERT model, with only a few differences. Based on the results of a grid search, the pretraining learning rate was set much higher than what was used to train the original ALBERT models, at 5.28e-3. Additionally, the input length was gradually increased during pretraining, following the curriculum learning approach proposed by \newcite{nagatsuka_cl_2021}. The training data used was identical to that used to train the FinBERT model and the same uncased tokenizer of FinBERT was also used for the FinALBERT models.

In the following section, we evaluate the FinSQuAD dataset, the MT method used, as well as the performance of the trained QA models.


\section{Evaluation}\label{sec4}

One of the main challenges with machine translated datasets is the absence of a large-enough, manually annotated, representative test set. Such a test set is in many cases difficult to create, as it entails replicating the entire annotation task and procedure, which is a major undertaking for tasks with complex annotations, such as QA. Therefore, in addition to reporting model performance on the machine translated test set, we also carry out several other evaluations: a backtranslation experiment, a manual evaluation of the translated examples, and a comparison of our method respective to two other machine translated SQuAD datasets, one for Spanish and one for Finnish.  These comparisons allow us in particular to establish the relative merits of our approach to other methods of SQuAD machine translation. 

We use as metrics the \emph{exact match (EM)}, the proportion of questions that receive the exactly correct answer span, and \emph{token F1}, the F1 score of the precision and recall of tokens in the predicted answer span, compared to the reference answer span. The latter metric is more tolerant to minor changes at the span boundaries.

\subsection{QA Model performance}

In Table~\ref{tab:otherlangs}, we compare the scores of our model to scores reported for other machine translated QA datasets. Our Finnish QA scores are the highest among those reported, well within the range that is to be expected with similar datasets. Of particular interest is the very substantial gain compared to the results \newcite{kylliainen_question_nodate} reported on the previously available Finnish translation of SQuAD2.0 but otherwise using a very comparable model. We will return to these results when discussing the relative merits of the machine translation methods later in Section~\ref{eval:othertransfer}.

For further comparison, we trained an English model based on a comparable pre-trained language model (BERT-base). This model reaches EM 74.2 and F1 77.6 on the original SQuAD2.0 data. The observed drop of 6.0pp EM and 3.9pp F1 is a combined effect of, at least, (a) noise introduced during translation and (b) any possible effect of the target language being Finnish, rather than English.

To our disappointment, the results also indicate that the models based on FinALBERT are not notably better than the models based on FinBERT, i.e.\ we were unable to replicate on Finnish the improvements in QA performance reported for English with the ALBERT model architecture. We also note that the results of the Finnish models are more closely grouped in general compared to the SQuAD results presented by \newcite{lan_albert_2020}. Further investigation is needed to ascertain whether this difference is due to the quality and amount of pre-training data used by the Finnish models, or something else entirely.

In the remainder of the Evaluation section, we turn our attention towards other means of evaluating our FinSQuAD dataset, as well as the MT method applied to produce it.

\begin{table*}[th]
  \centering
  \begin{tabular}{ll|rrr}
    
    \hline
    {\bf Model}  & {\bf Language and dataset} & {\bf EM} & {\bf F1} & {\bf Reported in}\\\hline
    BERT-base  & Finnish SQuAD2.0 (ours)  & 68.2 & 73.7 & this work \\
    BERT-large  & Finnish SQuAD2.0 (ours)  & 70.0 & 76.1 & this work \\
    ALBERT-xlarge & Finnish SQuAD2.0 (ours)  & 70.2 & 75.9 & this work\\\hline
    BERT-base & Finnish SQuAD2.0 [1] & 55.5 & 61.9 & [1] \\
    BERT-base & Spanish SQuAD2.0 [2] & 63.4 & 70.2  & online [3] \\
    BERT-base & Swedish SQuAD2.0 [4] & 66.7 & 70.1 & online [4] \\
    BERT-base & Indonesian SQuAD2.0 [5] & 51.6 & 69.1 & online [6] \\
    BERT-base & Persian ParSQuAD [7] & 62.4 & 65.3 & [7] \\
    \hline
    BERT-base & English SQuAD2.0 & 74.2 & 77.6 & this work\\\hline
  \end{tabular}
  \caption{Exact match (EM) and F1 scores of our models as well as scores reported for other machine translated SQuAD datasets as well as the original English SQuAD2.0. Citation list: [1] \newcite{kylliainen_question_nodate}, [2] ~\cite{carrino_automatic_nodate}, [3] ~\cite{spanishsquad}, [4] ~\cite{squad_v2_sv}, [5] ~\cite{indobert}, [6] ~\cite{indosquad}, [7] ~\cite{abadani_parsquad_2021}}
  \label{tab:otherlangs}

\end{table*}
\subsection{Evaluation through backtranslation}

The relative ease, with which the annotation transfer method can be applied to any language pair supported by the machine translation service, allows for a backtranslation-based evaluation. Here we translate our FinSQuAD data back to English, including the annotation transfer as if Finnish was the original language and English the target language. The resulting backtranslated English SQuAD dataset therefore accumulates errors over two rounds of translation, and can serve to estimate the impact on trained models due to errors incurred during the translation and annotation transfer. 

In Table~\ref{tab:backtranslate}, we report model performance measured on the original English SQuAD2.0 test set, comparing a model trained on the original English training data, with a model trained on the backtranslated training data. We see a drop of 8.4pp in terms of exact match, and 5.1pp in terms of F1. Considering that these are the result of two cumulative translation and annotation transfer rounds, we can expect the loss incurred on the Finnish model, after one round of translation, to be less. If the errors were to be assumed as approximately evenly distributed between the two rounds of translation, the negative impact would be around 4.2pp EM and 2.6pp F1. This can be seen as a rather acceptable ``price'' for a dataset obtained without any manual annotation.

\begin{table}[th]
  \centering
  \begin{tabular}{l|rr}
    \hline
    & \multicolumn{1}{c}{EM} & \multicolumn{1}{c}{F1} \\
    \hline
    Original         & 74.2 & 77.6 \\
    Backtranslated       & 65.8 & 72.5 \\
    \hline
  \end{tabular}
  \caption{Exact match (EM) and F1 scores between the original English SQuAD2.0 dataset and eng-fin-eng translated English dataset.}
  \label{tab:backtranslate}
\end{table}

\subsection{Evaluation respective to other transfer methods}\label{eval:othertransfer}

Direct comparison of the relative merits of our MT service -based annotation transfer method to its alternatives listed in Section~\ref{sec:relwork} is challenging, as these methods are very tedious to implement and replicate for new languages.

Nevertheless, a direct comparison is possible to the Finnish SQuAD2.0 dataset by \newcite{kylliainen_neural_2022}, which can be seen as an alternative translation of SQuAD2.0 to Finnish using a best-effort implementation of the translate-and-align approach. In all respects comparable QA models obtain F1 of 73.7 on our dataset compared to F1 of 61.9 on the dataset by \newcite{kylliainen_neural_2022}. Further, our translation loses 2.7\% of the original question-answer pairs in the process, compared to 28.1\% lost in the other dataset. These results seem to suggest that the translation method we used produces data of superior quality compared to the translate-and-align approach.

As a second point of comparison, we choose the Spanish QA dataset (as it has the highest reported scores after ours in Table~\ref{tab:otherlangs}, and can serve as a very strong baseline). The annotation transfer methods used to construct this dataset rely on language-specific resources and a technically complex pipeline, making a replication of the transfer method on Finnish tedious at best. Instead, we create a Spanish translation of SQuAD using our MT service-based method. We then train QA models on these two Spanish datasets using the Spanish ALBERT-XXL model\footnote{\url{https://huggingface.co/dccuchile/albert-xxlarge-spanish}}, and compare their relative performance. The results of this comparison are reported in Table~\ref{tab:spanish}. When trained and tested on the same dataset, the result seen earlier for the Finnish dataset repeats, here with a 3.7pp EM and 5.5pp F1 improvement in favor of our method of translation and annotation transfer. In cross-dataset experiments, we see that training on our dataset always brings better F1 score, irrespective of which test set we use. The EM metric then has an opposite tendency, hinting at the two methods producing different entity boundaries, which are then learned by the QA models.

\begin{table}[th]
  \centering
  \begin{tabular}{|ll|rr|}
    \hline
    Train & Test & \multicolumn{1}{c}{EM} & \multicolumn{1}{c|}{F1} \\
    \hline
    TAR   & TAR     & 66.3 & 73.7 \\
    our & TAR     & 64.5 & 74.0 \\
    TAR   & our   & 65.2 & 76.1 \\
    our & our   & 70.0 & 79.2 \\
    \hline
  \end{tabular}
  \caption{Exact match (EM) and F1 scores between Spanish TAR method and Spanish DeepL method.}
  \label{tab:spanish}
\end{table}

\subsection{Dataset error analysis}


Finally, we conducted a manual error analysis on a randomly selected subset of the FinSQuAD dataset, sampling 321 answerable questions from 51 passages in 17 different articles and inspected the resulting answer spans. We categorize the answers in 6 different categories:

\begin{tabular}{p{0.25\linewidth} p{0.60\linewidth}}
\textbf{Correct} & The answer span corresponded to the English original flawlessly \\
\textbf{Punctuation} & The answer corresponded to the English original, except for a minor difference in punctuation\\
\textbf{Over-extended} & The answer was longer than in the English original\\
\textbf{Under-extended} & The answer was shorter than in the English original\\
\textbf{Wrong} & The answer did not correspond to the English original of reasons other than over/under-extension.\\
\textbf{Missing} & The question did not have an answer, the span failed to be transferred\\
\end{tabular}

The result of the error analysis in Table~\ref{tab:dataset_eval} show that full 87.2\% of the answers are transferred fully correctly, and only 2.2\% of the answers are lost, i.e.\ not transferred at all. The most common error, accounting for nearly all errors in the dataset is over-extension, most typically by a single token.

\begin{table}[th]
  \centering
  \begin{tabular}{l|rr}
    \hline
    & \multicolumn{1}{c}{\#} & \multicolumn{1}{c}{\%} \\
    \hline
    Correct         & 280 & 87.2 \\
    Punctuation     & 1   & 0.3 \\
    Over-extended   & 29  & 9.0 \\
    Under-extended  & 4   & 1.2 \\
    Wrong           & 0   & 0.0 \\
    Missing         & 7   & 2.2 \\\hline
    Total           & 321 & 100.0 \\
    \hline
  \end{tabular}
  \caption{Error analysis results of the translated FinSQuAD dataset.}
  \label{tab:dataset_eval}
\end{table}

\section{Discussion and Conclusions}\label{sec5}

In this paper, we have demonstrated a practical method for annotation transfer through an affordable, high-quality machine translation service, relying on its ability to translate formatted text documents. We have applied this method to create a Finnish QA dataset with very little effort and negligible cost, resulting in a Finnish SQuAD2.0 translation with higher coverage and better overall model performance than what was previously available for Finnish. As a side product of our evaluation, we have also created an alternate Spanish SQuAD dataset of seemingly better quality than that previously available. We have shown, through comparison to other machine translated QA datasets, and more directly also through an English-Finnish-English backtranslation experiment, that the dataset is unlikely to result in substantially worse models than a (hypothetical) Finnish dataset created manually. The backtranslation experiment suggest the penalty for MT is about 5pp in terms of EM and 2.5pp in terms of F1.

We argue that the value of the approach is in allowing for a substantial expansion in the availability of numerous NLP tasks in a number of languages that currently lack the relevant native datasets. While it is clear that a high-quality dataset manually annotated in the target language is the best resource for training NLP models, it is clear that for many task-language pairs such a dataset will not be created for many years to come, if ever. In these cases, we argue that the method gives a practical, viable alternative which, thanks to its simplicity can be implemented with ease and applied quite broadly to produce datasets for many tasks in many languages. The applicability of the method will naturally depend on the task, and likely to a degree the language at hand.

The method is naturally limited to the language pairs supported by the translation service used and may not be practical for very large datasets in the billion word range. It also relies on the availability of a suitable translation service with terms and conditions not restricting such application (as is the case at present). While such dependence is not ideal, it is nevertheless becoming somewhat the norm in NLP, where large, high-quality models and systems are increasingly exposed through a service, rather than distributed openly, which is understandable given their development and deployment costs.

Our code is available under an open source license, and can be used to generate QA datasets for other languages supported by the translation service. The Finnish dataset is on the HuggingFace dataset repository as TurkuNLP/squad\_v2\_fi, the code is on GitHub as TurkuNLP/squad2-fi, and the Finnish model is on the HuggingFace model repository as TurkuNLP/bert-base-finnish-cased-squad2.

\section{Acknowledgments}

We thank Jenna Kanerva for fruitful discussions at the beginning of the study. Computational resources were provided by CSC - IT Centre of Science, Finland.
The research was supported by the Research Council of Finland funding.



\section{Limitations}

One limitation of this work is in relying on a particular property of an existing MT system, which also limits the applicability only to the languages supported by it. This is alleviated by the fact that DeepL supports 33 languages, allowing for a potentially very large number of datasets to be translated in the simple manner we outline. Further, since professional MT systems are primarily targeting translators and need to support formatting transfer to remain competitive, it is conceivable that a suitable MT system can be found also for other languages.

Another limitation is in relying on a closed, commercial system, which naturally negatively affects e.g.\ replicability. However, the system only needs to be used once, when creating the new dataset, and after that the dataset is available openly and can be evaluated in a transparent manner. The closed nature of the MT system thus does not fully transfer onto the dataset. We note that our use of a closed MT system is fully comparable to the current wide-spread practice in which NLP datasets are created using closed, commercial LLMs such as OpenAI's ChatGPT.

\bibliographystyle{acl_natbib}
\bibliography{nodalida2025}

\end{document}